\documentclass[runningheads]{llncs}
\usepackage[T1]{fontenc}
\usepackage{tikz}
\usepackage{caption}   
\usepackage{subcaption}

\usepackage{graphicx,verbatim}
\usepackage{multirow}
\usepackage{xcolor, colortbl}
\usepackage{makecell}
\usepackage{amsmath}
\usepackage{lineno}

\usepackage{booktabs}  
\usepackage{colortbl}  
\usepackage{xcolor}    
\usepackage{amsfonts}

\usepackage[table]{xcolor}
\providecommand{\STAB}[1]{\begin{tabular}{@{}c@{}}#1\end{tabular}}

\definecolor{lightgray}{gray}{0.94}

\usepackage[colorlinks=true, 
     linkcolor=blue, 
     filecolor=blue, 
     citecolor = blue,       
     urlcolor=blue]{hyperref}
\begin{document}
\title{\textit{ENC-ODE}: Event-level Neurodegenerative Modeling in Continuous Time with Neural ODEs}

\titlerunning{Event-level Neurodegenerative Modeling in Continuous Time}

\author{
Yujee Song\inst{1}$^\ast$
\and
Seunghun Baek\inst{1}$^\ast$
\and
Guorong Wu\inst{2}
\and
Won Hwa Kim\inst{1}
} 

\authorrunning{Y. Song \& S. Baek et al.}

\institute{
Pohang University of Science and Technology, Pohang, South Korea \\
\email{\{yujees, habaek4, wonhwa\}@postech.ac.kr}
\and
University of North Carolina at Chapel Hill, Chapel Hill, USA
\\
\email{guorong\_wu@med.unc.edu}
}

\def\thefootnote{*}\footnotetext{Y. Song and S. Baek contributed equally to this paper.}

\maketitle              
\begin{abstract}
\sloppy
Accurately predicting the temporal evolution of clinical biomarkers is crucial 
for the early diagnosis and management of neurodegenerative diseases such as Alzheimer’s disease.
However, this relies on longitudinal data to capture biomarker changes over time, which is often sparse and irregular due to the high cost, labor-intensive nature, and patient burden. 
To address these challenges, we propose ENC-ODE, 
an {\bf E}vent-level {\bf N}eurodegenerative modeling in {\bf C}ontinuous time with neural 
{\bf O}rdinary {\bf D}ifferential {\bf E}quations.
ENC-ODE predicts future biomarker evolution by modeling clinical events through diagnosis-conditioned continuous dynamics. 
A target-conditioned attention mechanism weights and aggregates event-level predictions for the target time and modality without history compression.
Extensive experiments on Alzheimer’s Disease Neuroimaging Initiative (ADNI) dataset 
demonstrate that ENC-ODE outperforms representative sequence models 
while offering a scalable and neuroscientifically grounded solution for clinical support.

\textit{The code is available at https://github.com/JardinDelSol/enc-ode.}

\end{abstract}

\section{Introduction}

Accurately modeling temporal dynamics of human brain is essential for improving patient management and early diagnosis of neurodegenerative diseases \cite{goenka2021deep,baek2023learning}.
For example, 
predicting how biomarkers such as Standardized Uptake Value Ratio (SUVR) of $\beta$-amyloid, metabolism, and Tau protein evolve over time 
helps clinicians 
to identify 
early symptoms of Alzheimer's disease (AD) and tailored treatment planning \cite{blennow2018biomarkers,cohen2014early,suvr}.
However, it is difficult to obtain a reliable prediction, as longitudinal neurobiological biomarkers are inherently sparse due to the high cost and patient burden from imaging scans~\cite{baek2024modality,baek2024ocl}. 
Thus, developing novel methods to predict future biomarker states from sparsely observed data is critical, as it provides a cost-effective alternative to radiation-based examinations.

Recent efforts explore diverse sequence modeling 
for neurodegenerative disease forecasting,  
which generally follow two primary paradigms categorized into discrete-time and continuous-time architectures.
Discrete-time architectures including RNN~\cite{rnn}, Transformer~\cite{transformer}, and Mamba~\cite{mamba} are designed for regular intervals. 
As a result, they struggle to capture the continuous dynamics of disease progression under irregular time gaps.
Alternatively, continuous-time models such as Neural Ordinary Differential Equations (ODEs)~\cite{chen2018neural}, ODE-RNN~\cite{odernn}, Neural Flow~\cite{neuralflow}, and Continuous Recurrent Unit (CRU)~\cite{cru} 
handle temporal irregularity. 
Despite their different approaches to time, both paradigms share a structural limitation that they predominantly compress entire multimodal histories into a single summarized latent representation. 
Such compression fails to capture explicit dependencies between specific past {\em events} (e.g., individual modality-specific biomarker observations acquired at irregular time points, along with the corresponding diagnostic stage) and the target time or modality. 
This structural bottleneck hinders learning how specific target predictions should selectively draw information from highly heterogeneous past observations,
which ultimately limits the precise characterization of long-term disease trajectories.

To overcome these limitations, we propose {\bf E}vent-level {\bf N}eurodegenerative {\bf C}ontinuous-time modeling with neural {\bf ODE}s (ENC-ODE).
ENC-ODE propagates each past observation through diagnosis-conditioned ODE dynamics and aggregates the resulting predictions using attention conditioned on the target time point and modality. This design explicitly models biomarker evolution under varying time intervals while avoiding the loss of event level information caused by compressing multimodal histories into a single latent state.
By computing attention scores 
on unobserved target event
and dynamically propagated predictions, 
our model prioritizes meaningful information 
over restrictive history compression.
This design effectively handles irregularly sampled data 
and provides 
precise characterization of diagnosis-specific biomarker trajectories. 


\sloppy
Our {\bf contributions} are summarized as follows:
\textbf{i)} {\em Diagnosis-conditioned event-level continuous trajectory modeling:} We propose diagnosis-driven dynamics for each individual observed event to capture stage-specific progression rates across irregular time intervals, enabling the precise characterization of continuous biomarker evolution.
\textbf{ii)} {\em Target-conditioned multimodal aggregation:} 
We introduce an attention module that selectively aggregates past observations based on their direct relevance to the target time and modality, bypassing the information loss inherent in previous unified latent bottlenecks.
\textbf{iii)} {\em Extensive validation:} We demonstrate the effectiveness of ENC-ODE via 
extensive analyses, validating its 
generation of plausible biomarker trajectories (i.e., changes of 
SUVR measures
) on Alzheimer’s Disease Neuroimaging Initiative (ADNI) dataset.

\section{Methods}

    
    
    

\begin{figure}[!t]
    \centering
    \includegraphics[width=1\textwidth]{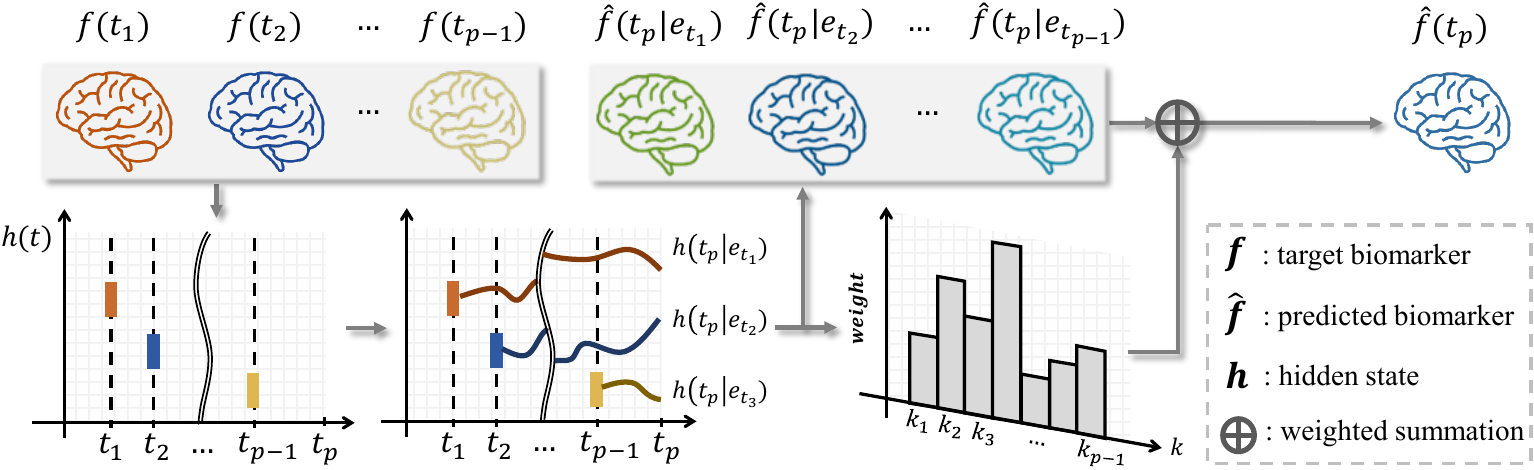}
    \caption{\small {\bf Overview of ENC-ODE.} 
    The input consists of multimodal brain regional measures $f(t_i)$ collected at irregular time points $t_i$. 
    The $f(t_i)$ is first encoded into a hidden state 
    $h$ where Neural ODEs model the continuous evolution of hidden states $\hat h$. 
    The learned trajectories capture biomarker dynamics, enabling the prediction of future states. 
    The final prediction $\hat{f}(t_p)$ aggregates event-level predictions via weighted summation, 
    reflecting the relative contribution of each prior event to the target prediction.
    }
    \label{fig:main}
\end{figure}

Our approach aims to model longitudinal recordings on multimodal brain imaging and predict the temporal evolution of the brain with respect to disease stages. 
We denote the 
longitudinal data as $\mathcal{H}_{t_n} := \{ e_{t_i} \} _{i=1} ^n$, where $t_1 < t_2 < \cdots < t_n$ are ordered time points and $e_i$ represents an event (e.g., image acquisition) 
at time $t_i$. 
Each event $e_{t_i} := \{ m_{t_i}, f_{t_i}, \delta_{t_i}, t_i\}$ consists of the modality type $m_{t_i}$ (e.g., 
SUVR of $\beta$-amyloid), 
the measurement 
from 
$N$ 
regions of interests (ROIs) 
$f_{t_i} \in \mathbb{R}^N$, 
the 
diagnosed 
AD stages $\delta_{t_i}$, 
and the observed time point $t_i$.

Similar to 
RNN and Transformer-based models, 
we first encode $e_{t_i}$ into a hidden state $h(t | e_{t_i}) \in \mathbb{R} ^{dim_h}$, but 
we disentangle event-driven changes over time using 
ODEs as shown in Fig. \ref{fig:main}.
Each observation captures different aspects of the subject's condition, and ODEs are used to model how the 
ROIs evolve based on the patient's state and diagnosis.
The predictions for the follow-up measurement, $\hat{f}(t_p)$, are made 
by solving the initial value problem~\cite{ivp} (IVP) and aggregating the resulting hidden states that incorporate prior events $\{ e_{t_i} \} _{i=1} ^{p-1}$.

\subsection{Hidden State Propagation}
As we consider observations made from different modalities, we first map our observations into latent representations. Specifically, the event $e_{t_i}$ is encoded into hidden state, $h(t_i|e_{t_i})$, using a neural network $g_{enc}(\cdot)$:
\begin{equation}
    h(t_i | e_{t_i}) = g_{enc} (e_{t_i}) = g_{enc}(\big[f_{t_i} \; || \; \xi_{m}(m_{t_i}) \; || \; \xi_{\delta}
    (\delta_{t_i})\big]),
\end{equation}
where $\xi_{m}(\cdot): \mathbb{R}\to\mathbb{R}^{dim_m}$ and $\xi_{\delta}(\cdot): \mathbb{R}\to\mathbb{R}^{dim_\delta}$ 
are learnable mappings that transform modality and diagnosis into $dim_m$ and $dim_\delta$ dimensional vectors. 


Brain measurements exhibit natural temporal fluctuations driven by aging and disease progression.
Therefore, the hidden state representing the patient's condition should also evolve over time. 
To address this,
our approach captures these evolutions by modeling the hidden state dynamics using an ODE:
{\begin{align}
dh(t | e_{t_i} ) &= \gamma \big(h(t | e_{t_i}), \xi_\delta(\delta_{t_i}); \theta \big) dt
\label{eq:odesolve}
\\
h(t_p| e_{t_i}) &= h(t_i|e_{t_i}) + \int _{t_i} ^{t_p} \gamma\big(h(s | e_{t_i}), \xi_\delta(\delta_{t_i}) ; \theta\big)ds
\end{align}}%
where $\gamma(\cdot)$ is a Neural ODE module parameterized by a neural network $\theta$. 
The function $\gamma(\cdot)$ models the evolution of the patient's state over time while incorporating diagnosis information $\delta$, enabling the model to learn diagnosis-conditioned dynamics, as individuals in different disease stage may exhibit 
heterogeneous dynamics. 
By solving the IVP of this ODE, we obtain the hidden state trajectory, which captures how the target hidden state $h(\cdot)$ evolves over time.
Neural ODEs 
effectively handle
sparse and irregular time points \cite{chen2018neural,odernn}, 
making them 
suitable 
for modeling hidden state dynamics between irregular observations.

Notably, $\gamma(\cdot)$ modeling the rate of change takes only its current state $h(s|e_{t_i})$ and the diagnosis $\xi_\delta(\delta_{t_i})$ as input. 
This implies that the entire process is conditioned solely on the initial event $e_{t_i}$ 
independent from other observations.
The independent propagation of hidden states makes the model flexible as 
it allows each trajectory to evolve without being constrained by previous or future observations. 
This differentiates our method from other approaches that focus on finding a representation summarizing past observation based on inter-event interactions, allowing us to assess the impact of each event on the predicted measurements in the future $t_p$. Further details are discussed in Sec. \ref{exp:attention}.
%

\noindent\textbf{Efficient Computation.}
Despite all the benefits of ODEs, their autoregressive nature may lead to high computational costs. 
To 
mitigate
this issue, we model the rate of change as a multidimensional system of ODEs:
{
\begin{align}
{d \over dt} \mathbf{h}(t_p) =
{d \over dt}
\begin{bmatrix}
h(t_p|e_{t_1}) \\
\vdots \\
h(t_p|e_{t_{p-1}})
\end{bmatrix}
=
\begin{bmatrix}
\gamma(h(t_p|e_{t_1}), \xi_\delta(\delta_{t_1})) \\
\vdots \\
\gamma(h(t_p|e_{t_{p-1}}), \xi_\delta(\delta_{t_{p-1}}))
\end{bmatrix}.
\label{eq:hidden parallel}
\end{align}}%
Eq. \eqref{eq:hidden parallel} computes rate of change of 
$h(t_p|e_{t_i})$ at time $t_p$ in parallel per $e_{t_{i}}$, reducing the complexity of handling a
sequence with $n$ observations 
from $O(n^2)$ to $O(n)$. 



\subsection{Estimation on Multi-modal Biomarkers from Prior Observations}
Given a propagated hidden state $h(t_p | e_{t_i})$, we can obtain a prediction on the 
measurements $f({t_p})$ at time $t_p$ given the event $e_{t_i}$ as follows:
\begin{equation}
\hat{f}(t_p | e_{t_i}) = g_{dec}(h(t_p| e_{t_i})),
\end{equation}
where $g_{dec}: \mathbb{R}^{dim_h} \to \mathbb{R}^{N}$ is the decoder network that maps 
the hidden state into the measurements of ROIs. 
Moreover, to ensure the most reliable estimation, all predictions from prior observations 
are aggregated. 
In our work, the final prediction $\hat{f}(t_p)$ is obtained by combining predictions from earlier observations $\{\hat{f}(t_p|e_{t_i})\}_{i=1}^{p-1}$, effectively capturing information from the entire sequence. 
To explicitly map the specific contribution of each past observation,
we aggregate all prior event-level 
trajectories through a weighted linear summation as
\begin{equation}
\hat{f}(t_p | \mathcal{H}_{t_{p-1}}) = \sum_{t_i < t_p} w_{i,p} \cdot \hat{f}(t_p|e_{t_i}),
\end{equation}
where $w_{i,p}$ represents the contribution of each past observation to ensure that 
$\hat{f}(t_p | \mathcal{H}_{t_{p-1}})$ effectively blends information across the entire sequence.

We leverage an attention mechanism to capture the complex relationships between the target prediction time and modality, allowing us to identify the most influential factors.
The value of each $w_{:,p}$ is calculated as follows:
\begin{equation}
w_{:,p} = \text{softmax} \left( Q([\mathbf{h}(t_p) || \xi_t (t_{1:p-1}) ] ) K( [\xi_t(t_p) || \xi_m(m_p)]) ^\top \over \sqrt{\text{dim}_h} \right),
\end{equation}
where $\mathbf{h}(\cdot)$ is a multidimensional hidden state from Eq.\eqref{eq:hidden parallel}, and \( Q(\cdot) \) and \( K(\cdot) \) are single-layer linear models that generate the query and key matrices, respectively. Instead of employing a traditional self-attention mechanism, which is commonly used for next-event prediction, we compute attention between our predicted values and the target time and modality. This approach enables the model to emphasize predictions \( \hat{f}(t_p | e_{t_i}) \) that are most relevant for estimating \( e_{t_p} \), making it particularly effective for handling irregularly sampled time points \cite{yang2022transformer}. The resulting attention weights provide insights into inter-modality relationships, enhancing 
cross-modal synergies,
as further discussed in Section~\ref{exp:attention}.

\begin{figure}[b]
    \centering
    \parbox{.64\textwidth}{
    \renewcommand{\arraystretch}{1.4}
    \centering
    \captionof{table}{Sample-size of the ADNI 
    dataset categorized by disease labels 
    based on 
    the first record.}
    \scalebox{0.67}{
    \renewcommand{\arraystretch}{1.08}
    \renewcommand{\tabcolsep}{0.16cm}
    \begin{tabular}{c|l||ccccc|c}
        \Xhline{2.5\arrayrulewidth}
        \textbf{Biomarker} & \multicolumn{1}{c||}{\textbf{Category}} & \textbf{CN} & \textbf{SMC} & \textbf{EMCI} & \textbf{LMCI} & \textbf{AD} & \textbf{Total} \\
        \hline
        \multirow{2}{*}{\textbf{FDG}} 
        & \# of subjects  & 255 & 4  & 189  & 277  & 115 & 840\\
        & \# of records   & 782 & 4  & 455  & 1016  & 607 & 2864\\
        \hline
        \multirow{2}{*}{\textbf{TAU}} 
        & \# of subjects  & 60  & 47  & 49  & 31  & 21 & 208\\
        & \# of records   & 140  & 112  & 122 & 76 & 56 & 506\\
        \hline
        \multirow{2}{*}{\textbf{AMY}} 
        & \# of subjects  & 220 & 89  & 237  & 158  & 56 & 760\\
        & \# of records   & 672 & 240 & 741  & 379  & 274 & 2306\\
        \Xhline{2.5\arrayrulewidth}
    \end{tabular}
    }
    \label{tab:biomarker}}
    \hfill
    \begin{minipage}{0.34\textwidth} 
        \captionof{table}{Averaged length of sequence in 
        both
        scenarios.}
        \parbox{.84\linewidth}{
        \renewcommand{\arraystretch}{1.72}
        \centering
        \scalebox{0.72}{
        \begin{tabular}{c | c c} \Xhline{0.25ex}
        \textbf{Biomarker} & \textbf{Unimodal} & \textbf{Multimodal} \\
        \hline
        \textbf{FDG} & 3.42 & 5.98\\
        \hline
        \textbf{TAU} & 2.43 & 7.31\\
        \hline
        \textbf{AMY} & 3.03 & 6.06\\
        \Xhline{0.25ex}      
        \end{tabular}
        \label{tab:sequence_length}}
        }
    \end{minipage}
\end{figure}

\section{Experiment and Results}


\subsection{Experiment Setup}

\noindent\textbf{Dataset.}\label{sec:dataset}
\sloppy
We use ADNI data~\cite{adni} to validate our framework.
Region-specific 
Standardized Uptake Value Ratio (SUVR)~\cite{suvr} of $\beta$-amyloid protein (AMY), metabolism (FDG), and Tau protein (TAU) from positron emission tomography (PET) scans are obtained at $N$$=$$160$ brain regions based on Destrieux atlas~\cite{destrieux}.
The dataset includes 
5676 records from 1176 subjects, detailed in Tab.~\ref{tab:biomarker}. 
Modality $m_t$ consists of $\{AMY, FDG, TAU\}$, 
and diagnosis $\delta_t$ represents the ordinal AD progression as Cognitive Normal (CN), Subjective Memory Complaint (SMC), Early Mild Cognitive Impairment (EMCI), Late MCI (LMCI) and AD.
Events are grouped by subjects and ordered chronologically.

\noindent\textbf{Implementation Details.\label{sec:implementation}}
We use a
fully connected layer (FCL)
for $g_{enc}$ and $g_{dec}$.
The ODE solver \cite{chen2018neural} uses the Euler method~\cite{euler_method} with a step size of $4$, 
and its
structure consists of two FCLs with activation normalization~\cite{actnorm}.
The model is trained using
root mean squared error (RMSE).
We train the model for 1000 epochs using 
Adam~\cite{adam} optimizer with a learning rate of 1e-4, a batch size of 1024, and a weight decay of 0.05.

\noindent\textbf{Event Prediction Settings.}
\label{sec:sequence_setting}
Given a sequence of observations made on continuous trajectory $f(t)$ for $t \in [t_1, t_{p})$, our goal is to predict $f(t_p|\mathcal{H}_{t_{p-1}})$.
Since observations occur at discrete time points, we optimize the model based on discrete data. 
During the training, the objective simplifies to predicting the final event $f(t_p)$ from the preceding $p-1$ events.
We conduct {\bf 1)} unimodal and {\bf 2)} multimodal sequence predictions to 
evaluate the contribution of individual biomarkers and the benefits of multimodal integration for 
contextual disease progression modeling.
In the unimodal setting, sequences include only one target modality (e.g., AMY) with a minimum length of 2. 
In the multimodal setting, sequences contain at least one instance of the target modality before the final observation, which is also the target modality, 
while non-target modalities can appear at any position elongating the sequence 
as summarized in Tab.~\ref{tab:sequence_length}.

To evaluate our method in capturing the temporal dynamics of Alzheimer's biomarkers, 
we compare its performance against representative sequence prediction models such as RNN~\cite{rnn}, Transformer~\cite{transformer}, Neural ODE~\cite{chen2018neural}, ODE-RNN~\cite{odernn}, Neural Flow~\cite{neuralflow}, CRU~\cite{cru} and Mamba~\cite{mamba}.
For fair comparisons, we designed baselines to have similar model capacity (i.e., the number of parameters). Also, we used the same set of training hyperparameters described in Sec. \ref{sec:implementation} across all baselines, and applied them consistently across different modalities. 
Root mean squared error (RMSE), mean absolute error (MAE) and coefficient of determination ($R^2$) were employed as evaluation metrics to quantify overall predictive performance. 
To ensure the robustness and reliability of our results, we replicated 
the same experiment five times and report the mean and standard deviation. 

\begin{table*}[t]
\caption{\small Comparison of Prediction Scores. Results are reported as the mean with the standard deviation over five independent runs. \textbf{Bold face} and \underline{underline} denote the best performance within each setting and overall, respectively.}
\centering
\renewcommand{\arraystretch}{1.5}
\renewcommand{\tabcolsep}{0.06cm}
\scalebox{0.522}{
\begin{tabular}{cl||ccc|ccc|ccc}
    \Xhline{4\arrayrulewidth}
    \multicolumn{2}{c||}{\textbf{Target}} & \multicolumn{3}{c|}{\textbf{FDG}} & \multicolumn{3}{c|}{\textbf{TAU}}& \multicolumn{3}{c}{\textbf{AMY}} \\
    \hline

    \multicolumn{2}{c||}{\textbf{Metric}}
    & \textbf{RMSE}$\,\downarrow$ & \textbf{MAE}$\,\downarrow$ & \textbf{$R^2\,\uparrow$} & \textbf{RMSE}$\,\downarrow$ & \textbf{MAE}$\,\downarrow$ & \textbf{$R^2\,\uparrow$} & \textbf{RMSE}$\,\downarrow$ & \textbf{MAE}$\,\downarrow$ & \textbf{$R^2\,\uparrow$}\\
    \Xhline{2.5\arrayrulewidth}

    \multirow{8}{*}{\STAB{\rotatebox[origin=c]{90}{Unimodal}}} & RNN
    & 0.1034$_{\pm0.0017}$ & 0.078$_{\pm0.0012}$ & 0.7739$_{\pm0.0072}$ & 0.2428$_{\pm0.0054}$ & 0.1338$_{\pm0.0031}$ & 0.7141$_{\pm0.0127}$& 0.2059$_{\pm0.0007}$ & 0.1561$_{\pm0.0005}$ & 0.7276$_{\pm0.0020}$ \\
    
    & Transformer
    & 0.1039$_{\pm0.0014}$ & 0.0779$_{\pm0.0008}$ & 0.7769$_{\pm0.0060}$ & 0.2225$_{\pm0.0041}$ & 0.1258$_{\pm0.0014}$ & 0.7599$_{\pm0.0088}$ & 0.1998$_{\pm0.0013}$ & 0.1506$_{\pm0.0010}$ & 0.7433$_{\pm0.0034}$\\

    & Neural ODE
    & 0.1068$_{\pm0.0013}$ & 0.0706$_{\pm0.0006}$ & 0.7642$_{\pm0.0058}$ & 0.1681$_{\pm0.0030}$ & 0.1105$_{\pm0.0020}$ & 0.8630$_{\pm0.0049}$ & 0.1897$_{\pm0.0010}$ & 0.1433$_{\pm0.0009}$ & 0.7687$_{\pm0.0025}$\\

    & ODE-RNN
    & 0.1027$_{\pm0.0010}$ & 0.0768$_{\pm0.0007}$ & 0.7827$_{\pm0.0041}$ & 0.2211$_{\pm0.0049}$ & 0.1287$_{\pm0.0022}$ & 0.7631$_{\pm0.0107}$ & 0.1901$_{\pm0.0023}$ & 0.1427$_{\pm0.0019}$ & 0.7677$_{\pm0.0057}$\\
    
    & Neural Flows
    & 0.1089$_{\pm0.0003}$ & 0.0703$_{\pm0.0002}$ & 0.7552$_{\pm0.0016}$ & 0.1613$_{\pm0.0036}$ & 0.1076$_{\pm0.0020}$ & 0.8738$_{\pm0.0056}$ & 0.1887$_{\pm0.0012}$ & 0.1418$_{\pm0.0010}$ & 0.7711$_{\pm0.0029}$\\

    & CRU
    & 0.1138$_{\pm0.0018}$ & 0.0866$_{\pm0.0016}$ & 0.7326$_{\pm0.0086}$ & 0.1913$_{\pm0.0040}$ & 0.1236$_{\pm0.0018}$ & 0.8226$_{\pm0.0075}$ & 0.2021$_{\pm0.0013}$ & 0.1551$_{\pm0.0011}$ & 0.7375$_{\pm0.0034}$\\
     
    & Mamba
    & 0.0964$_{\pm0.0017}$ & 0.0706$_{\pm0.0015}$ & 0.8082$_{\pm0.0067}$ & 0.1715$_{\pm0.0044}$ & 0.1096$_{\pm0.0029}$ & 0.8573$_{\pm0.0074}$ & 0.2006$_{\pm0.0056}$ & 0.1502$_{\pm0.0053}$ & 0.7412$_{\pm0.0142}$\\

    
    & \cellcolor{gray!20}ENC-ODE (ours)
    & \cellcolor{gray!20}\textbf{0.0882$_{\pm0.0023}$}
    & \cellcolor{gray!20}\textbf{0.0621$_{\pm0.0006}$}
    & \cellcolor{gray!20}\textbf{0.8435$_{\pm0.0044}$} 
    & \cellcolor{gray!20}\textbf{0.1443$_{\pm0.0013}$}
    & \cellcolor{gray!20}\textbf{0.0956$_{\pm0.0004}$}
    & \cellcolor{gray!20}\underline{\textbf{0.8991$_{\pm0.0017}$}} 
    & \cellcolor{gray!20}\textbf{0.1855$_{\pm0.0028}$}
    & \cellcolor{gray!20}\textbf{0.1394$_{\pm0.0021}$}
    & \cellcolor{gray!20}\textbf{0.7788$_{\pm0.0067}$} \\
    \Xhline{2.5\arrayrulewidth}

    \multirow{8}{*}{\STAB{\rotatebox[origin=c]{90}{Multimodal}}} & RNN
    & 0.1006$_{\pm0.0001}$ & 0.0768$_{\pm0.0001}$ & 0.7900$_{\pm0.0004}$ & 0.2401$_{\pm0.0030}$ & 0.1448$_{\pm0.0022}$ & 0.5367$_{\pm0.0122}$ & 0.2171$_{\pm0.0016}$ & 0.1597$_{\pm0.0009}$ & 0.6974$_{\pm0.0045}$\\

    & Transformer
    & 0.0984$_{\pm0.0010}$ & 0.0751$_{\pm0.0007}$ & 0.7991$_{\pm0.0037}$ & 0.2122$_{\pm0.0054}$ & 0.1271$_{\pm0.0027}$ & 0.6401$_{\pm0.0181}$& 0.1995$_{\pm0.0014}$ & 0.1503$_{\pm0.0007}$ & 0.7444$_{\pm0.0037}$\\

    & Neural ODE
    & 0.0914$_{\pm0.0004}$ & 0.0694$_{\pm0.0003}$ & 0.8265$_{\pm0.0014}$ & 0.2671$_{\pm0.0045}$ & 0.1668$_{\pm0.0029}$ & 0.4302$_{\pm0.0192}$& 0.2806$_{\pm0.0024}$ & 0.2108$_{\pm0.0019}$ & 0.4944$_{\pm0.0087}$\\

    & ODE-RNN
    & 0.0936$_{\pm0.0010}$ & 0.0715$_{\pm0.0008}$ & 0.8179$_{\pm0.0040}$ & 0.2276$_{\pm0.0023}$ & 0.1346$_{\pm0.0020}$ & 0.5862$_{\pm0.0085}$& 0.2391$_{\pm0.0037}$ & 0.1784$_{\pm0.0031}$ & 0.6330$_{\pm0.0115}$\\
    
    & Neural Flows
    & 0.1155$_{\pm0.0018}$ & 0.0890$_{\pm0.0014}$ & 0.7230$_{\pm0.0085}$ & 0.2742$_{\pm0.0052}$ & 0.1842$_{\pm0.0045}$ & 0.3995$_{\pm0.0227}$& 0.2234$_{\pm0.0024}$ & 0.1708$_{\pm0.0019}$ & 0.6795$_{\pm0.0069}$\\

    & CRU
    & 0.1157$_{\pm0.0024}$ & 0.0895$_{\pm0.0020}$ & 0.7217$_{\pm0.0117}$ & 0.2711$_{\pm0.0045}$ & 0.1836$_{\pm0.0033}$ & 0.4127$_{\pm0.0196}$& 0.2305$_{\pm0.0026}$ & 0.1742$_{\pm0.0027}$ & 0.6590$_{\pm0.0076}$\\
    & Mamba
    & 0.0959$_{\pm0.0021}$& 0.0729$_{\pm0.0015}$ & 0.8091$_{\pm0.0082}$& 0.2016$_{\pm0.0046}$ & 0.1275$_{\pm0.0035}$ & 0.6752$_{\pm0.0147}$& 0.2059$_{\pm0.0038}$ & 0.1536$_{\pm0.0021}$ & 0.7277$_{\pm0.0100}$\\


    & \cellcolor{gray!20}ENC-ODE (ours)
    & \cellcolor{gray!20}\underline{$\textbf{0.0760}_{\boldsymbol{\pm0.0006}}$}
    & \cellcolor{gray!20}\underline{$\textbf{0.0579}_{\boldsymbol{\pm0.0005}}$}
    & \cellcolor{gray!20}\underline{$\textbf{0.8802}_{\boldsymbol{\pm0.0019}}$}
    & \cellcolor{gray!20}\underline{$\textbf{0.1397}_{\boldsymbol{\pm0.0002}}$}
    & \cellcolor{gray!20}\underline{$\textbf{0.0921}_{\boldsymbol{\pm0.0005}}$}
    & \cellcolor{gray!20}$\textbf{0.8442}_{\boldsymbol{\pm0.0005}}$
    & \cellcolor{gray!20}\underline{$\textbf{0.1781}_{\boldsymbol{\pm0.0022}}$}
    & \cellcolor{gray!20}\underline{$\textbf{0.1351}_{\boldsymbol{\pm0.0019}}$}
    & \cellcolor{gray!20}\underline{$\textbf{0.7964}_{\boldsymbol{\pm0.0050}}$}\\
    
    \Xhline{3\arrayrulewidth}
\end{tabular}}
\label{tab:prediction_score}
\end{table*}

\subsection{Experimental Results and Analysis}

\noindent\textbf{Comparison on Prediction Score.} 
Tab.~\ref{tab:prediction_score} demonstrates that ENC-ODE outperforms baselines in both unimodal and multimodal settings, 
achieving the highest performance across all metrics.
In unimodal scenarios, its superior performance stems from the diagnosis-conditioned ODEs, which precisely characterize continuous biomarker evolution by accounting for stage-specific progression rates over irregular time intervals.
Notably, ENC-ODE gains performance from incorporating different modality types in prediction, 
whereas others rather deteriorated in many cases.
This highlights that target-aware attention effectively mitigates information interference by bypassing restrictive compression and selectively aggregating relevant features across disparate modalities.


\begin{figure}[!t]
    \centering
    \includegraphics[width=0.95\textwidth]{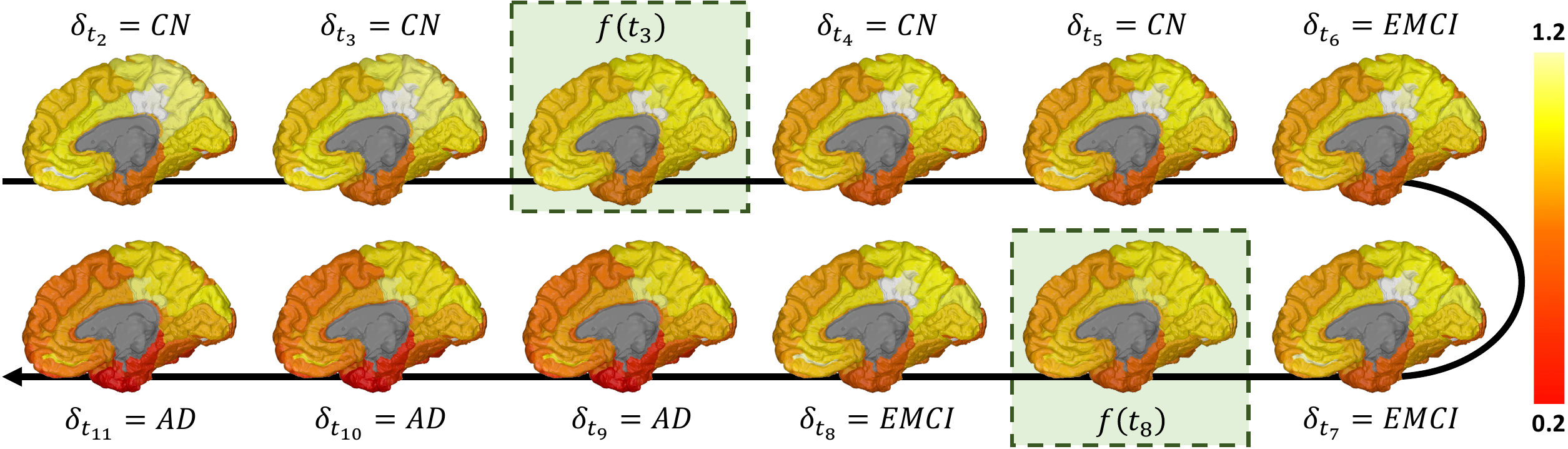}
    \caption{\small 
    Visualization of predicted FDG SUVR sequences (subject ID: 129\_S\_4422). The model captures the continuous decline in glucose metabolism as the disease progresses from CN to AD. 
    Green boxes denote actual FDG observations at $t_3$ and $t_8$.
    }
    \label{fig:traj}
\end{figure}

\noindent\textbf{Visualization of Actual Brain Trajectories.}
For a multimodal sequence of length 11 from a subject (ID: $129\_S\_4422$), Fig.~\ref{fig:traj} illustrates the predicted evolution of FDG SUVR over time. 
We include ground truth values at $t_3$ and $t_8$ to provide a direct comparison at the specific time points where FDG were measured.
As discussed in literature~\cite{FDG_decrease1,FDG_decrease2},
FDG generally decreases over time and shows a noticeable decline when there is a significant change in the diagnostic label (e.g., EMCI to AD at $t_9$).
Also, abrupt changes in parahippocampal gyrus and posterior cingulate 
show 
the clinical validity 
based on articles~\cite{parahippocampal,posterior_cingulate,overall_fdg}.
The close alignment between our predicted trajectories $\hat{f_t}$ and the ground truth measurements $f(t)$ at $t_3$ and $t_8$ further reinforces the reliability of our framework.

\begin{figure}[t]
    \centering
    \begin{minipage}{0.58\textwidth} 
        \captionof{table}{\small Ablation on diagnosis driven ODEs and attention aggregation using RMSE.
        }
        \renewcommand{\arraystretch}{1.15}
        \centering
        \scalebox{0.72}{
        \begin{tabular}{l | c | c | c} \Xhline{0.3ex}
        & \textbf{FDG} & \textbf{Tau} & \textbf{AMY} \\
        \hline
        w/o Diagnosis ODEs & 0.0816$_{\pm0.006}$ & 0.1511$_{\pm0.041}$ & 0.1818$_{\pm0.012}$ \\
        Uniform Aggregation   & 0.0918$_{\pm0.003}$ & 0.2315$_{\pm0.007}$ & 0.2114$_{\pm0.004}$ \\
        Latest Observation Only   & 0.0988$_{\pm0.002}$ & 0.2377$_{\pm0.023}$ & 0.2800$_{\pm0.005}$ \\
        \cellcolor{gray!20}ENC-ODE (Ours) & \cellcolor{gray!20}\textbf{0.0760$_{\pm0.006}$} & \cellcolor{gray!20}\textbf{0.1397$_{\pm0.001}$} & \cellcolor{gray!20}\textbf{0.1781$_{\pm0.002}$} \\
        \Xhline{0.3ex}
        \end{tabular}
        }
        \label{tab:ablation}
    \end{minipage}
    \hfill
    \parbox{.40\textwidth}{
    \renewcommand{\arraystretch}{1.638}
    \renewcommand{\tabcolsep}{0.18cm}
    \centering
    \captionof{table}{\footnotesize Average training time 
    per epoch with 
    and without 
    Eq.\eqref{eq:hidden parallel}.}
    \scalebox{0.78}{
    \begin{tabular}{l|c} \Xhline{0.3ex}
        Sequential \;(w/o Eq.\eqref{eq:hidden parallel}) & $25.66$s\quad \\
        \hline
        \rowcolor{gray!20}
        Parallel \;\;\;\; (w/ \ Eq.\eqref{eq:hidden parallel}) & $10.47$s \\
        \hline
        \hline
        Time Ratio & $40.8\%$ \\
    \Xhline{0.3ex}
    \end{tabular}
    \label{tab:parallel}
    }}
\end{figure}

\begin{figure}[t]
    \centering
    \begin{minipage}{0.35\textwidth} 
        \centering
        \includegraphics[width=0.9\linewidth, height = 2.7cm]{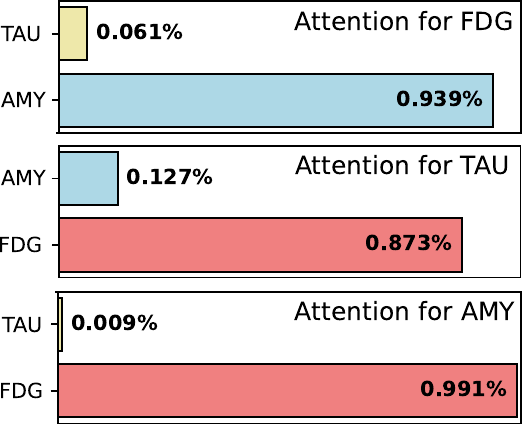}
        \label{fig:attention_score}
    \end{minipage}%
    \hfill
    \begin{minipage}{0.63\textwidth} 
        \centering
        \includegraphics[width=0.86\linewidth]{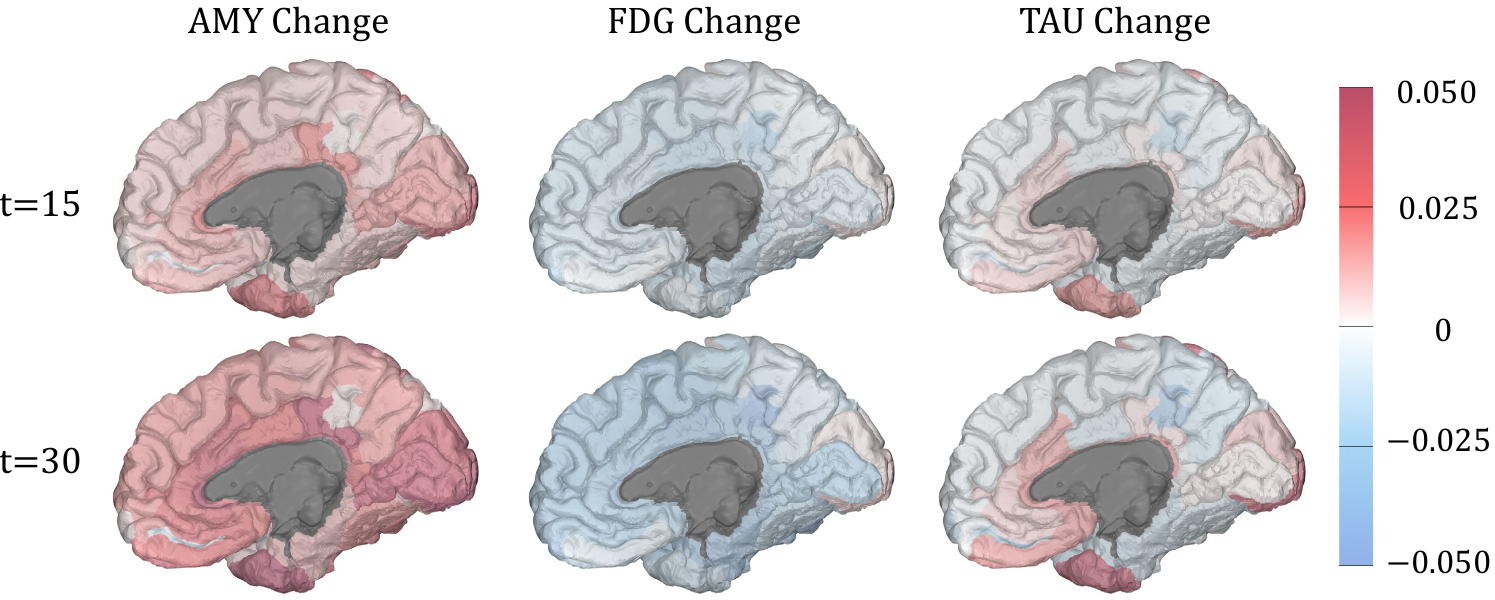}
        \label{fig:average_change}
    \end{minipage}
    \caption{\small Left: Attention Score from different modalities, Right: Averaged rate of change during $t \in [15, 30]$ in FDG prediction from all modalities $m_t \in \{AMY, FDG, TAU\}$.}
    \label{fig:attention}
\end{figure}
\noindent\textbf{Ablation on accuracy.} Tab.~\ref{tab:ablation} summarizes the impact of our proposed modules,
averaged over five different random seeds. 
To assess the influence of the diagnosis information in ODE modeling, we removed the diagnosis term $\xi_\delta(\delta_{t_i})$ from Eq.\eqref{eq:odesolve}
as follows: 
$dh(t | e_{t_i} ) = \gamma \big(h(t | e_{t_i}); \theta \big) dt$.
Furthermore, we evaluated the necessity of our attention mechanism by comparing it against uniform aggregation and a baseline using only the latest observation. 
The removal or simplification of each component led to a noticeable drop in performance across all settings, resulting in significantly higher RMSE values.
In particular, simplifying attention module showed significant degradation in Tau prediction, where Tau measurements account for only $8.9\%$ of the total observations (Tab.~\ref{tab:biomarker}). 
These results demonstrate that relying on a single recent event or simple averaging is insufficient to capture the complex dependencies within multimodal histories.

\noindent\textbf{Ablation on efficiency.} 
\label{exp:abl_eff}
Tab.~\ref{tab:parallel} highlights the efficiency 
of our parallel training strategy
(Eq.~\eqref{eq:hidden parallel}). 
Unlike 
a sequential approach
regarding 
each hidden state trajectory as an independent ODE, our method reduces the computational time,
including backpropagation, by 
$59.2\%$
, demonstrating its scalability and efficiency.

\noindent\textbf{Attention Weight Analysis.\label{exp:attention}}
The performance degradation under uniform aggregation or exclusive reliance on the latest observation (Tab.~\ref{tab:ablation}) underscores the necessity of our selective weighting strategy.
While it is straightforward that the attention mostly focuses on the same modality type,
to examine which modality provides complementary information in prediction,
we visualize the relative attention scores of two other modality types in Fig.~\ref{fig:attention} (left).
Given that FDG contributes significantly to the predictions, 
Fig.~\ref{fig:attention} (right) illustrates the temporal evolution of FDG-driven influences on each biomarker.
Notable changes in Amyloid and Tau stand out in 
regions including 
inferior temporal, cuneus, and posterior cingulate cortex,
which are established AD-specific regions~\cite{inferior_temporal,cunues,posterior_cingulate}.

\section{Conclusion}
\sloppy
In this work, we presented ENC-ODE, a continuous-time framework designed to navigate the temporal irregularity of multimodal neurodegenerative disease data. 
Our approach leverages diagnosis-conditioned dynamics to capture the nuanced, 
stage-specific tempo of brain biomarker evolution. 
In addition, the integration of target-aware attention enables a selective deconstruction of multimodal histories, 
effectively bypassing the information bottlenecks of traditional sequence models while providing a clear window into event-level influences. 
Validated on the ADNI dataset, ENC-ODE not only achieves superior predictive performance but also offers a clinically grounded and computationally efficient paradigm for longitudinal analysis. 
This work provides a high-resolution predictive paradigm that can support personalized treatment planning through event-level analysis.

\begin{credits}
\subsubsection{\ackname}
This research was supported by 
RS-2026-25494850 (60\%), 
RS-2025-02216257 (35\%), and
RS-2019-II1091906 (AI Graduate Program at POSTECH, 5\%).
\subsubsection{\discintname}
The authors have no competing interests to declare that are relevant to the content of this article.
\end{credits}


\bibliographystyle{splncs04}
\bibliography{Paper-1872}

\end{document}